\documentclass[letterpaper, 10 pt, conference]{ieeeconf}  

\IEEEoverridecommandlockouts                              

\title{\LARGE \bf
Soft Arm-Motor Thrust Characterization for a Pneumatically Actuated Soft Morphing Quadrotor
}

\author{Vidya Sumathy, Jakub Haluska, and George Nikolakopoulos
\thanks{This work is funded by European Union's Horizon Europe Research and Innovation Program, under the Grant Agreement No. 101119774 SPEAR.}
\thanks{The authors are with the Robotics \& AI Group, Department of Computer, Electrical and Space Engineering, Lule\r{a} University of Technology, Lule\r{a} SE-97187, Sweden. {Corresponding author: \tt\small vidya.sumathy@ltu.se}}
}

\usepackage{graphicx}
\usepackage{subcaption}
\usepackage{cite}
\usepackage{amsmath}
\usepackage{amssymb}
\usepackage{mathtools}

\begin{document}

\maketitle
\thispagestyle{empty}
\pagestyle{empty}

\begin{abstract}

In this work, an experimental characterization of the configuration space of a soft, pneumatically actuated morphing quadrotor is presented, with a focus on precise thrust characterization of its flexible arms, considering the effect of downwash. Unlike traditional quadrotors, the soft drone has pneumatically actuated arms, introducing complex, nonlinear interactions between motor thrust and arm deformation, which make precise control challenging. The silicone arms are actuated using differential pressure to achieve flexibility and thus have a variable workspace compared to their fixed counterparts. The deflection of the soft arms during compression and expansion is controlled throughout the flight. However, in real time, the downwash from the motor attached at the tip of the soft arm generates a significant and random disturbance on the arm. This disturbance affects both the desired deflection of the arm and the overall stability of the system. To address this factor, an experimental characterization of the effect of downwash on the deflection angle of the arm is conducted. 
\end{abstract}

\section{Introduction}
The 'soft' characteristic in soft aerial vehicles, as opposed to conventional rigid drones, is defined by various attributes such as the use of soft materials, soft actuators, and control systems that accommodate their flexible properties and often with bio-inspired design strategies \cite{tanaka2022review}.
Soft drones are extensively used in applications such as search and rescue \cite{fabris2021soft}, full-body perching on pipelines and irregular surfaces \cite{ruiz2022ieee, nguyen2024crash}, and safe interaction with delicate objects, such as grasping \cite{fishman2021dynamic, ubellacker2024high, cheung2024modular, guo2024powerful}. The write-up in \cite{floreano2017foldable} presents detailed examples of biologically inspired drones with folding mechanisms for morphological adaptation, size reduction, and mechanical resilience. While some works have addressed the implementation of soft actuators in drones, the current state-of-the-art has seldom explored the use and characterization of these soft actuators and provided an in-depth analysis of their behavior under the influence of downwash.

The Soft Morphing Quadrotor (SMQ) considered in this work is a custom-built soft drone equipped with four Soft Pneumatically Actuated (SPA) arms designed to withstand the vertical forces generated by the quadrotors’ motors while altering its morphology in flight. The soft arms deflect based on the differential pressure input, and the deflection is represented as an angle, referred to as the deflection angle, measured between the arm base and the servo motor. Since the deflection of the soft arms causes a change in the configuration of the drone, it is indispensable to consider the soft arm deflection as a system state. Correspondingly, the four differential pressure inputs are appended to the four control inputs of the drone. In soft drones, the downwash from the propellers attached at the tip of the flexible arms affects not only the stability of the drone but also the structure itself. As the deflection angle is one of the states of the soft drone, achieving stability and desired trajectory tracking requires counteracting or nullifying the effect of downwash on the deflection angle. Thus, the contributions of this paper are to empirically study the feasible structural configuration space of the soft drone as a function of the deflection angle of the soft arms and to experimentally characterize and analyze the effect of downwash on the deflection angle of the soft arms and its impact on the thrust vector.
\begin{figure}
\centering
    \begin{subfigure}{0.2\textwidth}
\includegraphics[scale=0.03]{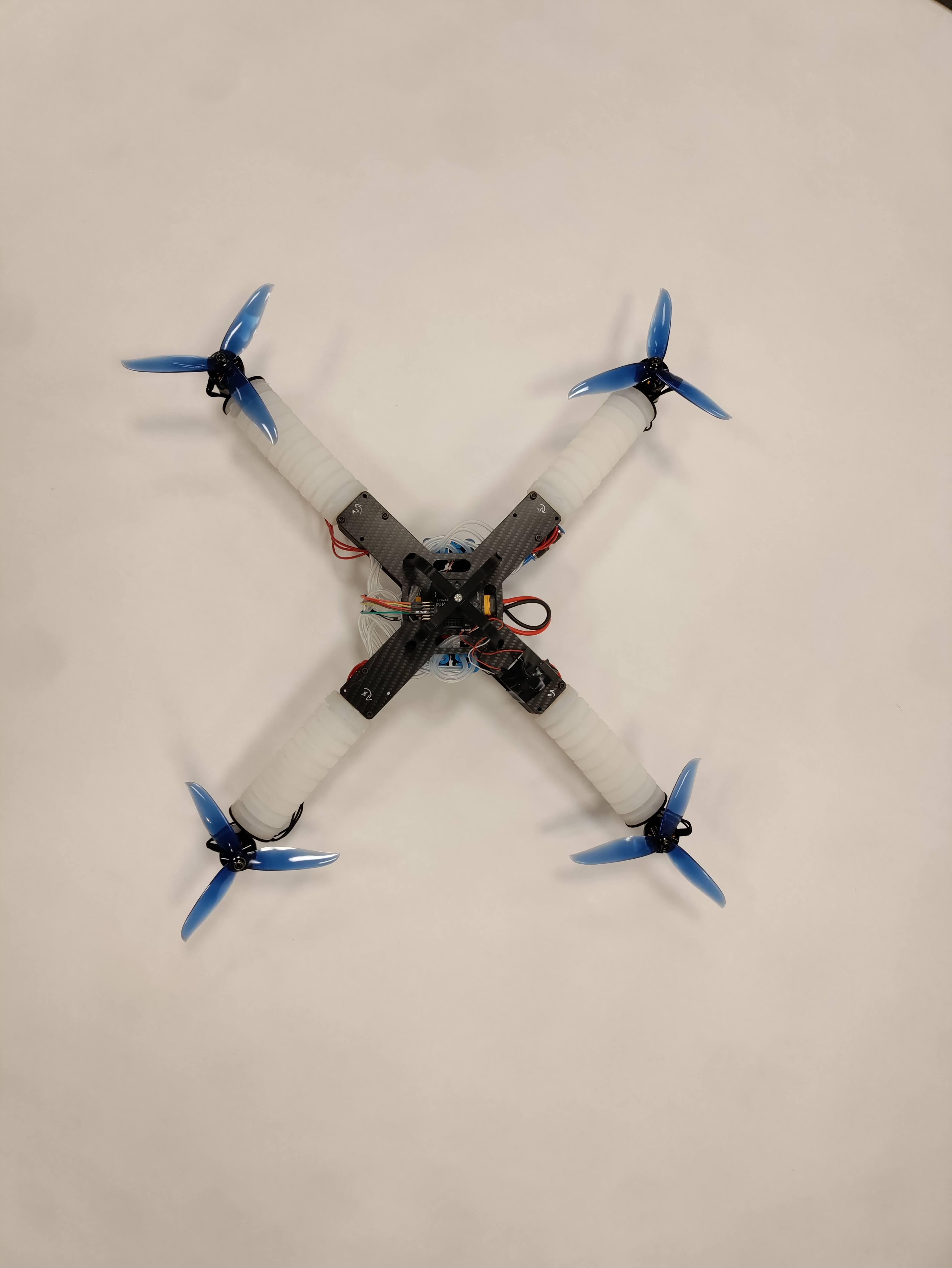}
    \end{subfigure}
    \begin{subfigure}{0.2\textwidth}
\includegraphics[scale=0.03]{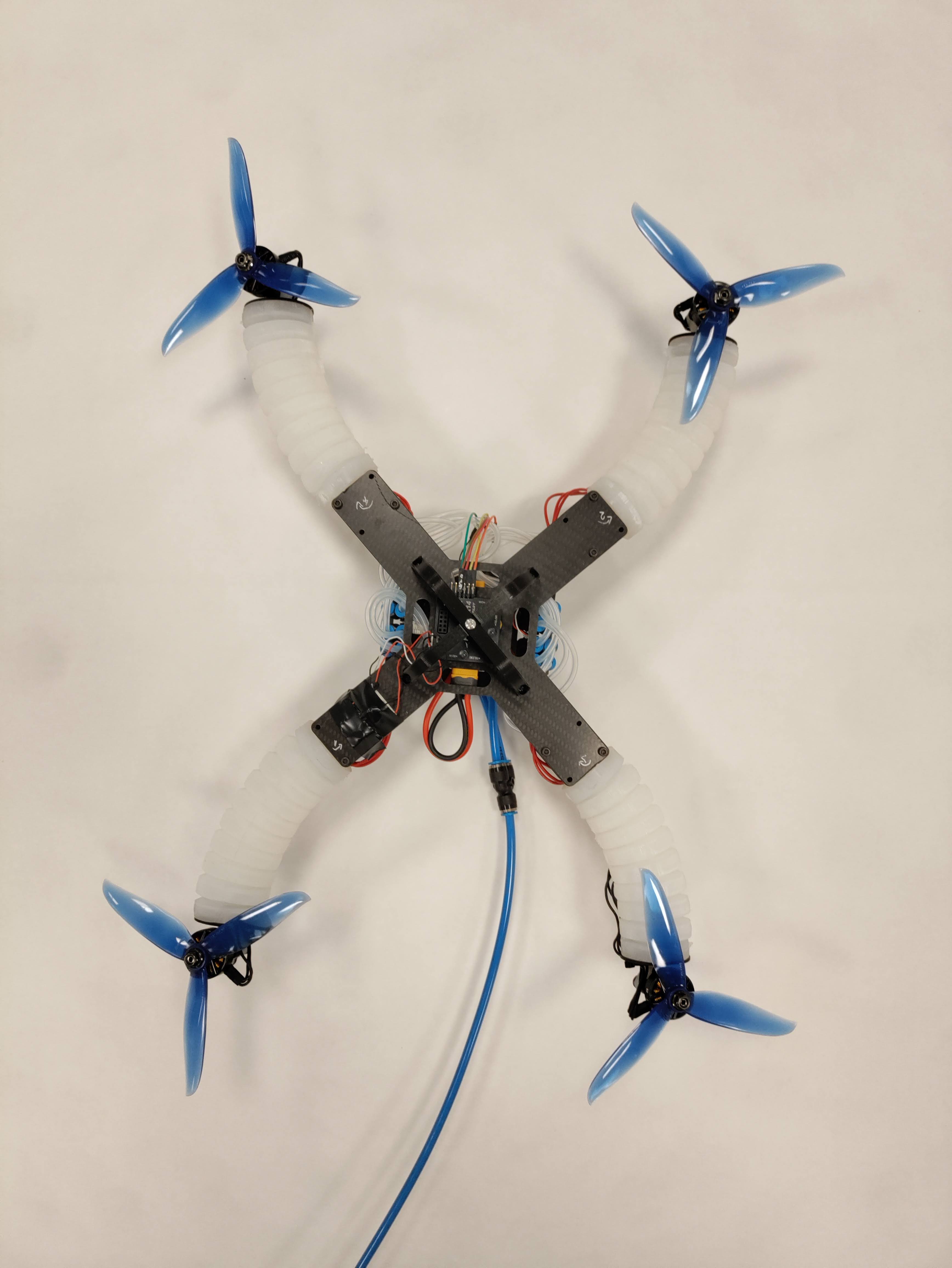}
    \end{subfigure}        
\caption{Soft drone with (a) unactuated soft arm and (b) pneumatically actuated soft arm.}
\label{fig_sd}
\end{figure}

\section{Soft Morphing Quadrotor}
The structure of the SMQ consists of four SPA arms controlled by pneumatic actuation and attached to the center via a carbon fiber frame. The soft drone, as shown in Fig. \ref{fig_sd}, features a Pixracer R15 flight controller, four T-motor F40PRO IV kV2400 motors for thrust equipped with T5150 3-blade propellers, a T-motor P50A 6S 4-IN-1 Electronic Speed Controller (ESC), and a ZIPPY Compact 1400 mAh 4S 65C LiPo battery, with a total weight of approximately 1 kg ~\cite{Haluska2022med}. The SPA arms have two main components: a semi-rigid inner spine printed using nylon filament on an FDM 3D printer and the active component, a silicone arm with air chambers that can be actuated with pressurized air. Each SPA arm comprises four chambers that are bidirectionally actuated to achieve positioning in the horizontal plane. The semi-rigid inner spines provide structural support against upward forces while allowing motion in the horizontal plane.

\section{Methodology}
\begin{figure}
\centering
    \begin{subfigure}{0.2\textwidth}
\includegraphics[scale=0.21]{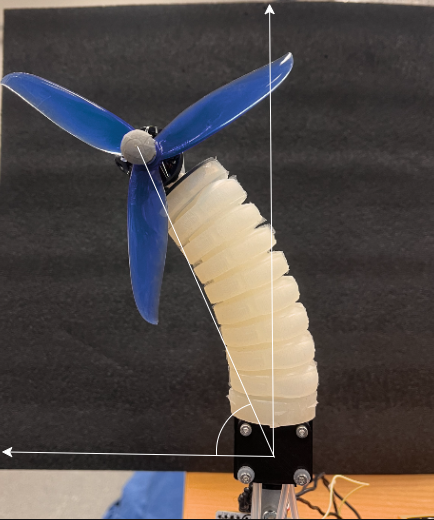}
    \end{subfigure}
    \begin{subfigure}{0.2\textwidth}
\includegraphics[scale=0.5]{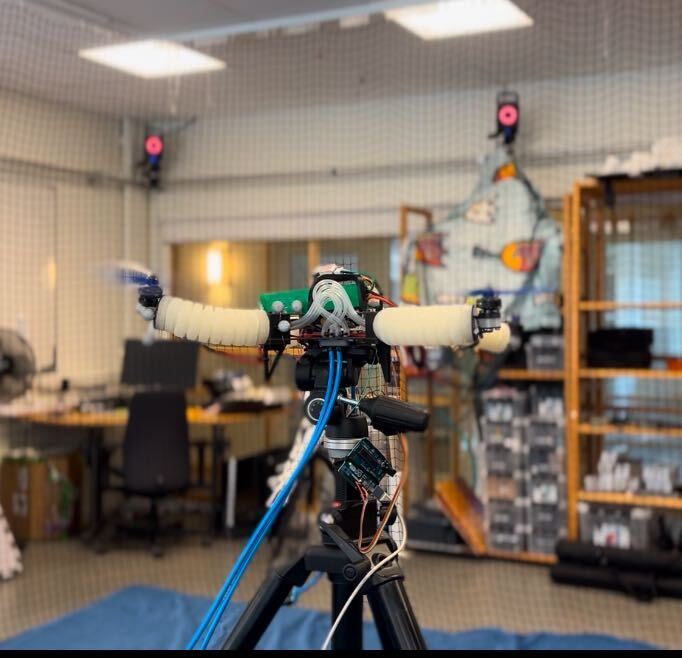}
    \end{subfigure}        
\caption{(a) The deflection angle of the soft arm with pneumatic actuation, and (b) experimental setup of the SMQ.}
\label{fig_exp}
\end{figure}

In the proposed work, the deflections of the four SPA arms are also considered as states to model the SMQ dynamics. The deflection is measured as the angle between the normal and deflected configurations of the arm, relative to its point of attachment to the center of the drone. This angle, referred to as the deflection angle $(\delta_{i})$, where $i \in {1,2,3,4}$, corresponds to each arm, as shown in Fig. \ref{fig_exp} (a). Thus, the state vector is defined as $q = [x,y,z,\phi,\theta,\psi,u,v,w,p,q,r,\delta_{1},\delta_{2},\delta_{3},\delta_{4}]$ where $q \in \mathrm{R^{16}}$ and the control input vector $u = [F, \tau_{\phi}, \tau_{\theta}, \tau_{\psi},dp_{1},dp_{2},dp_{3},dp_{4}]^{T}$ where $u \in \mathrm{R^{8}}$ and $dp_{i}$ is the differential pressure on each arm. The deflection of the arms introduces two noteworthy effects on the soft drone: first, a change in the configuration space of the drone, and second, the impact of downwash on the thrust vector.
The SMQ can achieve both 'X' and 'H' configurations based on the deflection of the soft arms. Since all four arms can deflect, two distinct 'H' configurations are possible. These changes in configuration make the computation of $u$ for achieving the desired pose more complex. This underscores the importance of including the deflection angle of each soft arm in the state vector $q$ of the system. During flight, the downwash from the motors and servo motors, attached to the tips of the soft drone's arms, causes an upward bend in the arms. This bend is empirically measured as angles $\Theta_{i}$ and $\Phi_{i}$ relative to the drone's vertical planes $XZ$ and $YZ$. As a result, only a component of the total thrust acts downward, leading to a considerable reduction in the thrust available for control. The remaining outward component of thrust generates a significant disturbance.
\section{Experiments} 
\begin{figure}
\centering
    \begin{subfigure}{0.2\textwidth}
\includegraphics[scale=0.07]{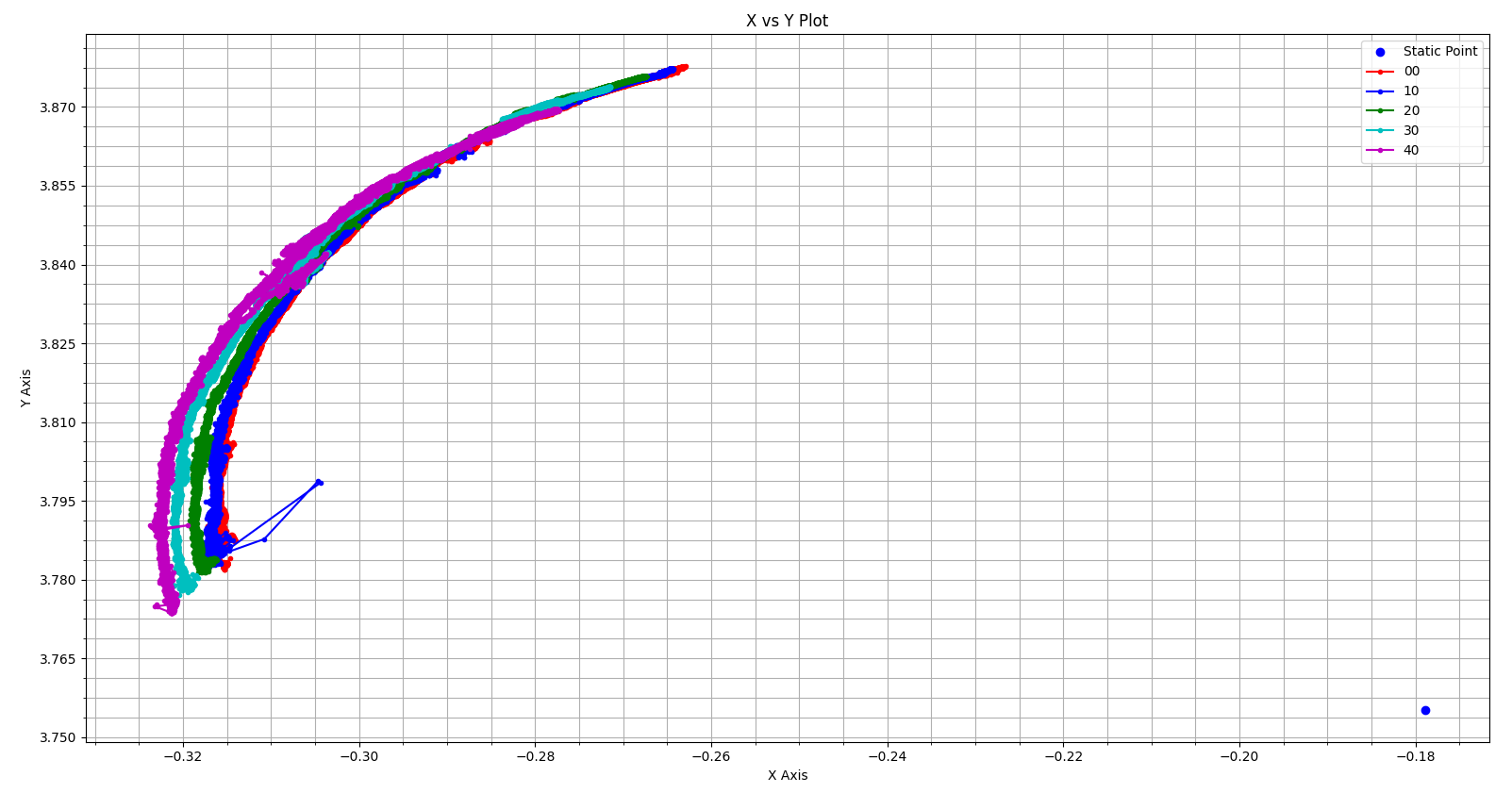}
    \end{subfigure}
    \begin{subfigure}{0.2\textwidth}
\includegraphics[scale=0.07]{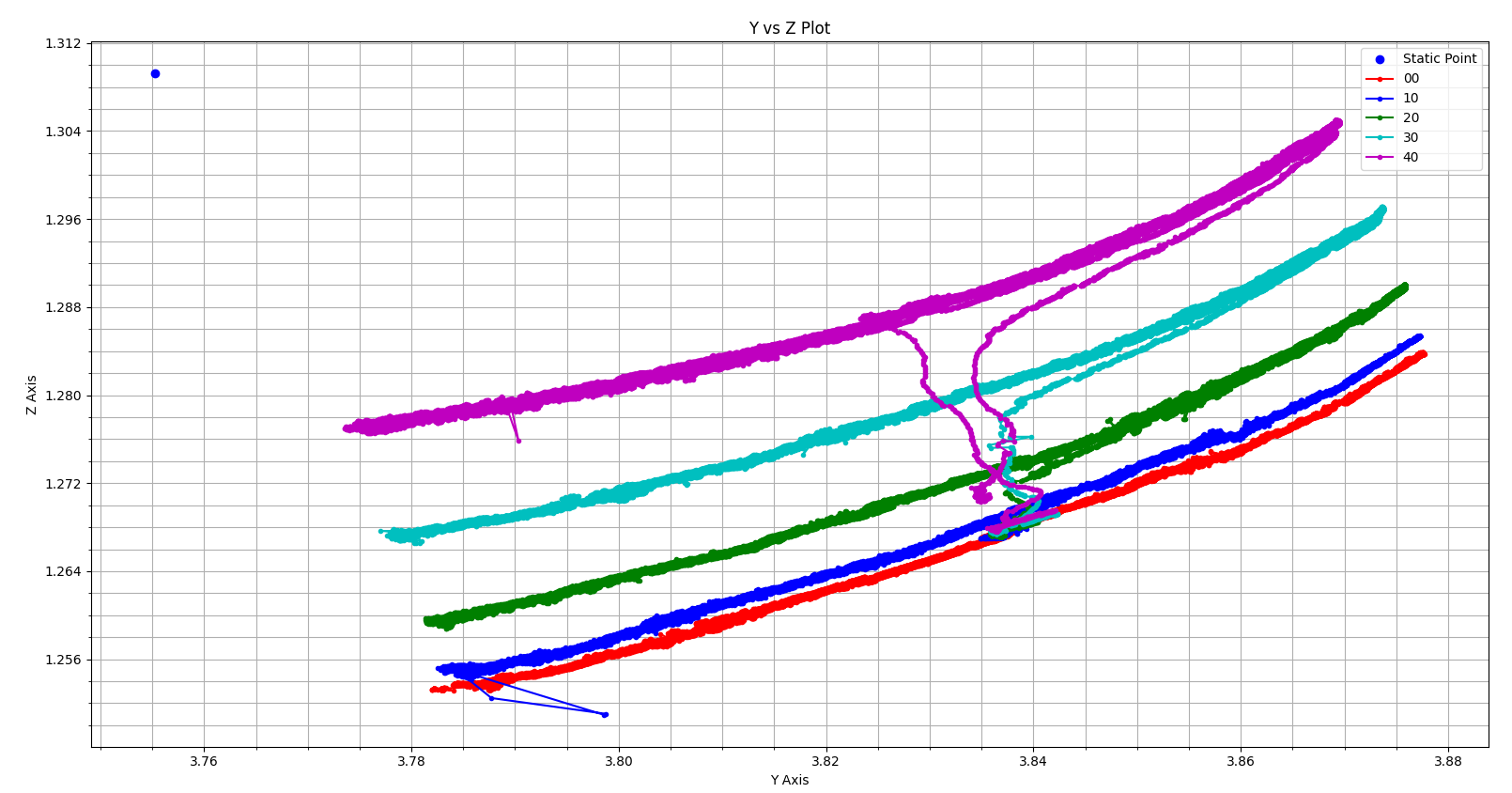}
    \end{subfigure}        
    \begin{subfigure}{0.2\textwidth}
\includegraphics[scale=0.07]{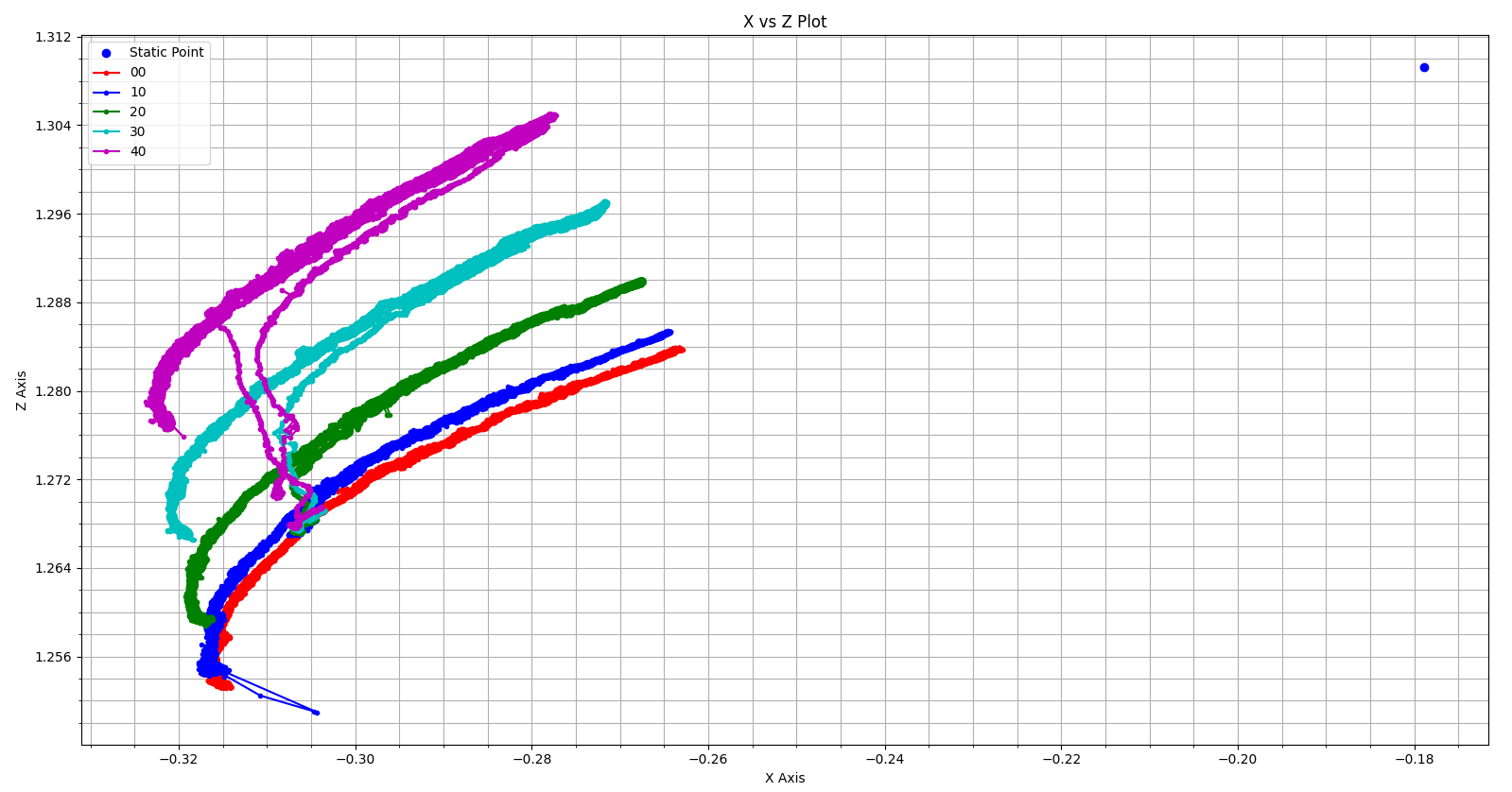}
    \end{subfigure}
    \begin{subfigure}{0.2\textwidth}
\includegraphics[scale=0.07]{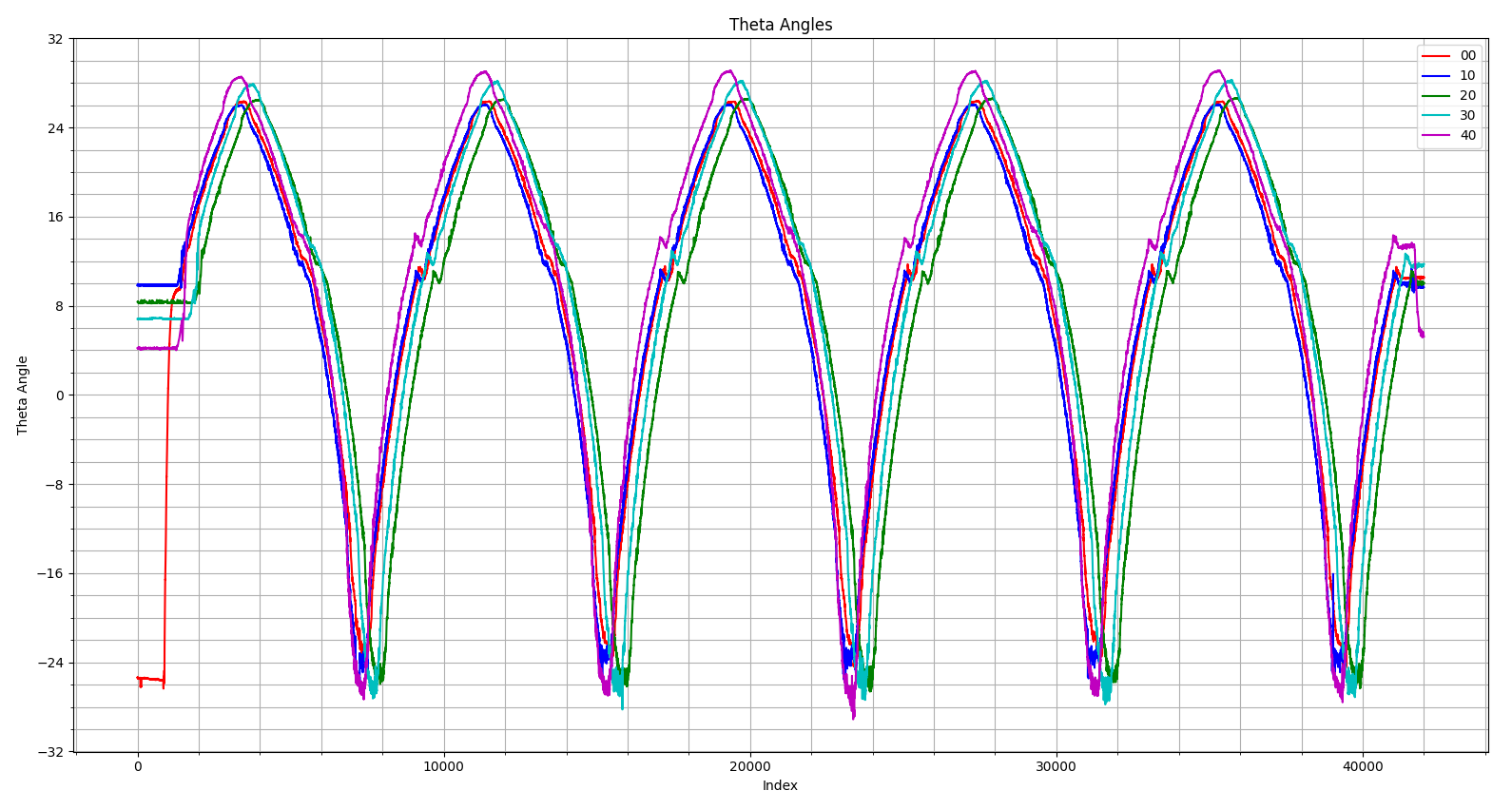}
    \end{subfigure}   
        \begin{subfigure}{0.2\textwidth}
\includegraphics[scale=0.07]{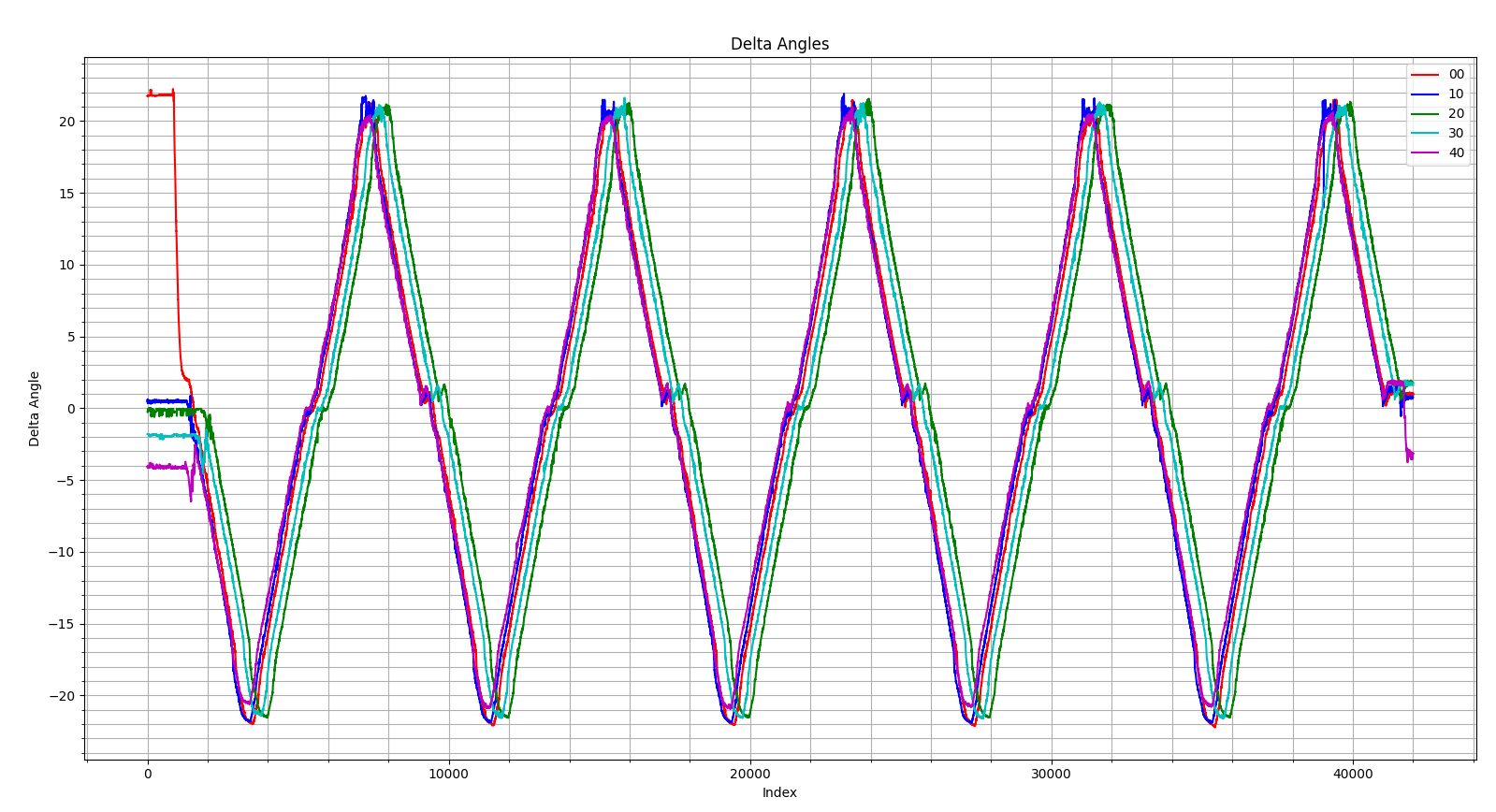}
    \end{subfigure} 
        \begin{subfigure}{0.2\textwidth}
\includegraphics[scale=0.07]{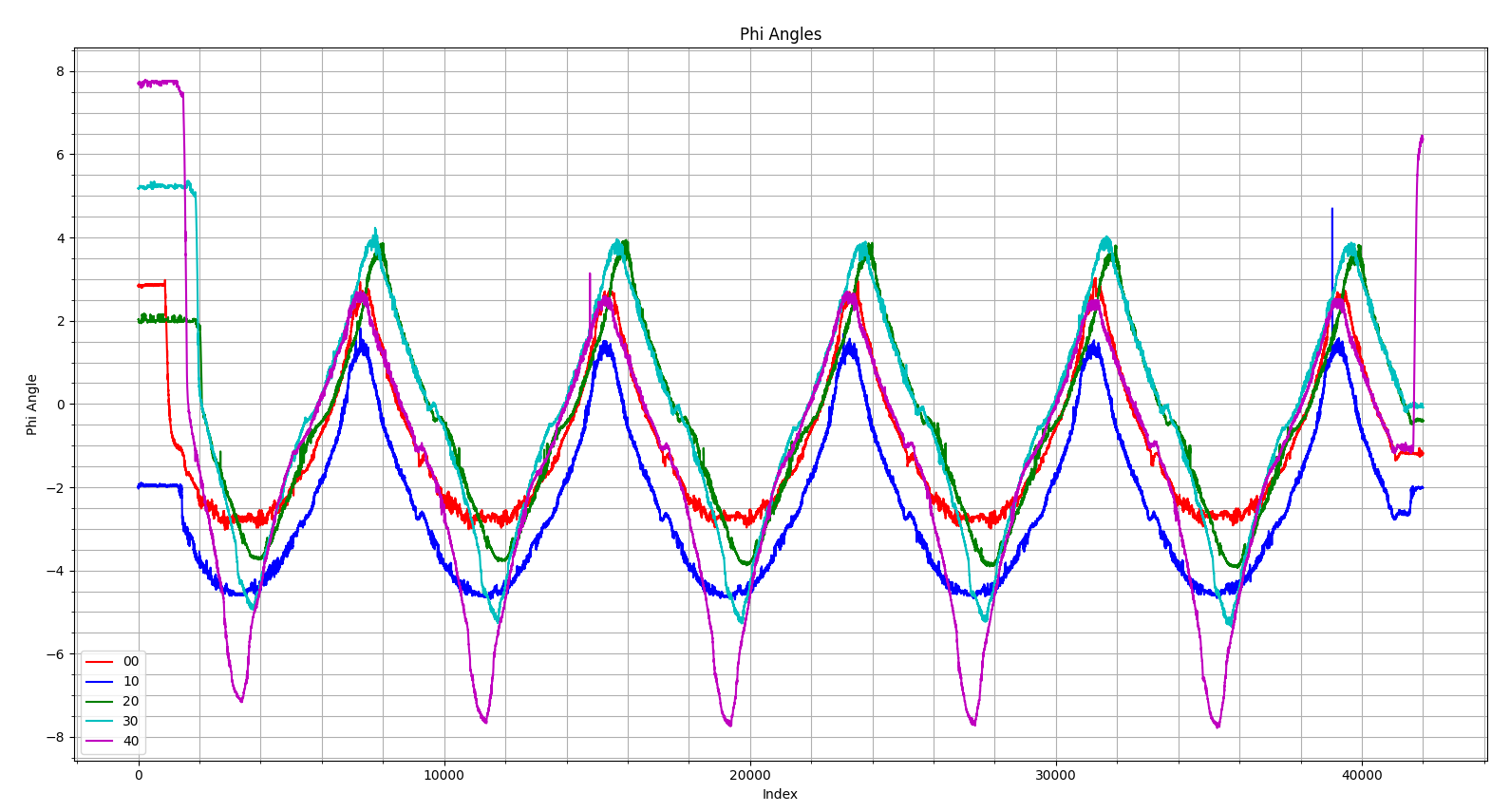}
    \end{subfigure} 
\caption{Experimental results show the motion of the tip of the SPA in (a) the horizontal plane $XY$, vertical planes (b) $YZ$ and (c) $XZ$, and changes in the (d) angle in the vertical plane $\Theta_{i}$, (e) deflection angle $\delta_{i}$, and (f) angle in the vertical plane $\Phi_{i}$, with and without motor thrust for the same $dp_{i}$}
\label{fig_results}
\end{figure}

The experimental setup, as shown in Fig. \ref{fig_exp} (b), consists of the SMQ, where the arm with the motor and propeller is attached to a thrust-measuring instrument called RC Benchmark to measure RPM, thrust, ESC power, and other parameters. The experiment is conducted on a single SPA of the drone. First, the soft arm is actuated without powering the propeller (0$\%$ thrust) to determine the undisturbed natural deflection angle $\delta_{i}$ of the soft arm. The differential pressure is varied from its minimum to maximum values, causing the arm to deflect in both directions. In the next set of experiments, the motors are powered while the arms are actuated to the same $dp_{i}$ as in the first experiment. The motor thrust is increased sequentially to 10$\%$, 20$\%$, 30$\%$ and 40$\%$, with each experiment repeated five times. This generates thrust and, correspondingly, downwash on the SPA while the arm is in motion. The deflection angle of the arm is measured using the Vicon system, which records the pose of the arm tip (where the motor is placed) and the base. The results, shown in Fig. \ref{fig_results} [a-c], depict the motion of the SPA tip in the horizontal plane $XY$ and vertical planes $YZ$ and $XZ$, respectively, with and without motor thrust for the same $dp_{i}$. The static point in the graph corresponds to the base of the arm. Figures \ref{fig_results} [d-f] show the corresponding change in the angle in the vertical plane $\Theta_{i}$, deflection angle $\delta_{i}$ of the arm, and the angle $\Phi_{i}$, respectively. As observed in the plots, as motor thrust increases, $\delta_{i}$ changes significantly. The components of motion in the vertical planes $XZ$ and $YZ$ and the corresponding values of $\Theta_{i}$ and $\Phi_{i}$ increase with thrust, highlighting the effect of downwash and vibration disturbances on the SPA deflection angle. The data collected from these experiments are then used to derive a mathematical relation between the deflection angle and the applied pressure, accounting for the effects of downwash and disturbances.

\section{Conclusion}
This study presents an experimental characterization of the configuration space of a soft morphing quadrotor with pneumatically actuated flexible arms. It evaluates the effects of motor downwash on the deflection angle of the soft arms and its subsequent influence on thrust vectoring and system stability. The results show the importance of the deflection angle as a system state and its differential pressure input as a system control parameter to ensure trajectory tracking and stability. Future work includes designing a hybrid data-driven MPC framework for the SMQ.
\bibliographystyle{vancouver}
\bibliography{soft_drone.bib}
\end{document}